\documentclass[lettersize,journal]{IEEEtran}
\usepackage{amsmath,amsfonts}
\usepackage{algorithmic}
\usepackage{algorithm}
\usepackage{array}
\usepackage[caption=false,font=normalsize,labelfont=sf,textfont=sf]{subfig}
\usepackage{textcomp}
\usepackage{stfloats}
\usepackage{siunitx}
\usepackage{url}
\usepackage{booktabs}
\usepackage{verbatim}
\usepackage{amsfonts,amsmath,amssymb,amsthm}
\usepackage{makecell}
\usepackage{graphicx}
\usepackage{multirow}
\usepackage{cite}
\newif\ifanonymous
\anonymousfalse

\begin{document}

\title{Better Calibrated, Better Planned: Uncertainty-Aware Multi-ASV Informative Path Planning for Oil Spill Monitoring}

\ifanonymous
  \author{Anonymous Authors}   
\else
  \author{Alejandro Casado-Pérez,
Marija Popović,
Daniel Gutiérrez-Reina,
Sergio L. Toral
and Samuel Yanes%
    \thanks{This work has been submitted to the IEEE for possible publication. Copyright may be transferred without notice, after which this version may no longer be accessible.}
    \thanks{This work was supported by Project  PID2024-158365OB-C21 funded by MICIU/AEI/10.13039/501100011033 and by FEDER, UE
And Universidad de Sevilla. (2022). VII Plan Propio de Investigación y Transferencia (2022-2025). Vicerrectorado de Investigación.}%
    \thanks{A. Casado-Pérez, D. Gutiérrez-Reina, S. L. Toral and S. Yanes are with the Departamento de Ingeniería Electrónica, Escuela Técnica Superior de Ingeniería, Universidad de Sevilla, 41005 Sevilla, España (email: acasado4@us.es)    }
    \thanks{M. Popovi\'c is with  MAVLab, Faculty of Aerospace Engineering, TU Delft, The Netherlands}}
\fi



\maketitle

\begin{abstract}
Monitoring oil spills with fleets of Autonomous Surface Vehicles (ASVs) requires mapping a heterogeneous pollutant field from sparse, noisy measurements under tight budgets. Informative Path Planning (IPP) couples a reconstruction model, which infers the field and its uncertainty, with a planner that decides where to sample next. Prior work has focused on the planner and largely relied on Gaussian Processes (GPs), leaving the calibration of the model's uncertainty underexplored. We introduce Evidential Deep Learning (EDL) as the information model of a fleet of ASVs sharing an uncertainty map, and benchmark it against Monte Carlo Dropout, Deep Ensembles and a GP in closed loop with five planners on physics-based oil spill simulations. Among the deep models, the model explains 80\% of the variance of the final error and the planner 1\%. EDL achieves the lowest error and the highest IoU under every planner, is the best calibrated model and runs at 3.9 ms per step. With 50 cells per vehicle, it outperforms MC Dropout and Deep Ensembles with 150. Deep Ensembles become more confident as their error grows, whereas EDL does not. Robustness to the data its own planner collects is therefore a requirement of uncertainty-aware IPP.

\end{abstract}

\section{Introduction}

Oil spills are one of the main threats to water bodies, and reducing pollutant discharges is an explicit target of Sustainable Development Goal 6 \cite{ods_guide}. Mapping a spill requires sampling a heterogeneous field over a large area with a few vehicles and a limited budget of time or energy.

Informative Path Planning (IPP)  ~\cite{POPOVIC2024104727} studies how a vehicle, or a fleet of vehicles, should move in order to maximize the information acquired about an environment under strict limits of budget, time, or energy. Rather than following a predefined route, each vehicle decides its next action as a function of the information it has already collected, so that sensing and planning become coupled processes. The measurements gathered by the fleet are processed into an information model that serves as a representation of the state of the environment and updates as new data arrive. This model turns the sparse measurements into an uncertainty map used by the planner to take intelligent decisions that optimize a defined task, be it information gathering, geofencing or detection of pollution sources.

IPP is therefore only as effective as its information model, however advanced the planner \cite{dutta2025ipp,fazliu2026xit,moon2026iatigris}.
If the uncertainty does not reflect where the model is wrong, adaptivity brings little benefit over a precomputed coverage path~\cite{chen2022ak}. 
The information model must meet three requirements at once. First, an accurate reconstruction which is the objective of the mission. Second, an uncertainty that locates the reconstruction error, which is the signal the planner acts upon. Third, a cost low enough to be queried at every decision step of every vehicle. Gaussian Processes (GPs) \cite{gaussianmachine} are the usual choice in IPP as they naturally provide a mean and uncertainty map, but their cost grows with the number of samples and their variance depends mainly on where the samples were taken rather than on what was measured. Deep uncertainty-aware models reconstruct the field at a constant cost, but sampling-based ones such as Monte Carlo Dropout (MC Dropout) \cite{pmlrv48gal16} and Deep Ensembles \cite{NIPS2017_9ef2ed4b} need several forward passes per prediction. Evidential Deep Learning (EDL)~\cite{NEURIPS2020_aab08546} returns the reconstruction and its aleatoric and epistemic uncertainty in a single forward pass.

\begin{figure*}[t]
  \centering
  \includegraphics[width=\textwidth]{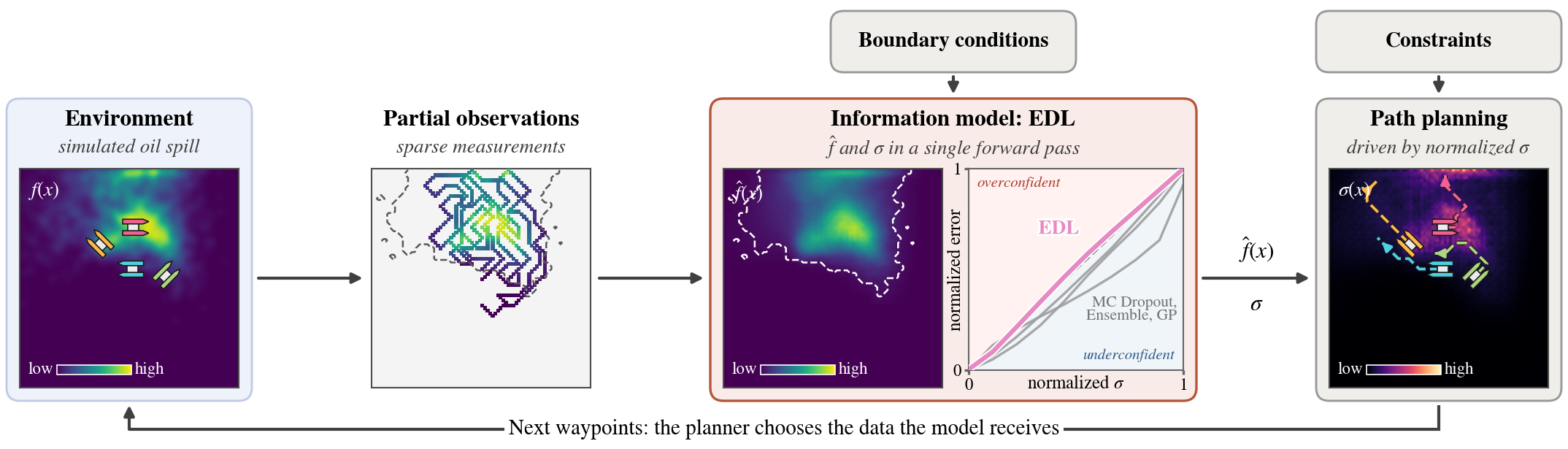}
  \caption{Uncertainty-aware multi-agent IPP loop studied in this paper, shown at one step of a simulated mission. A fleet of four ASVs samples a physics-based oil spill simulation $f(x)$. The sparse measurements are passed to the information model, EDL, which returns in a single forward pass the reconstruction $\hat{f}(x)$ (dashed: ground-truth spill contour) and its uncertainty $\sigma(x)$, subject to boundary conditions. The inset compares the error-uncertainty calibration of EDL with that of MC Dropout, Deep Ensembles and a GP. The planner uses the normalized $\sigma_{total}$ to choose the next waypoints of each agent (dashed arrows) under the mission constraints, and therefore chooses the data the model receives at the next step.}
  \label{fig:state_problem}
\end{figure*}

These three requirements are usually assessed offline, on data drawn from the training distribution. In IPP, however, the model operates inside a closed loop, illustrated in Fig. \ref{fig:state_problem}. The fleet collects sparse measurements of the field, the information model turns them into a reconstruction and an uncertainty map, and the planner uses that reconstructed uncertainty to choose the next waypoints of each vehicle, which determine the observations the model receives at the next step. This loop adds a fourth requirement that is easy to overlook. Because the planner decides what the model observes, in operation the model is queried on data whose distribution its own uncertainty has shaped, not on the data it was trained on. An uncertainty-guided fleet concentrates its samples where the uncertainty is high, typically on the spill (dashed arrows in Fig. \ref{fig:state_problem}), and a model that is accurate offline can fail on such data.
If its predictions are inaccurate or its uncertainty is miscalibrated, the planner is steered towards uninformative regions, and IPP degrades into blind exploration. Conversely, a model that predicts not only the value of the monitored variable but also quantifies the reliability of each prediction tells the planner where new measurements are most valuable.
We test the premise that better uncertainty estimates lead to better navigation and lower reconstruction error. Accordingly, we study how the calibration of the reconstruction model affects decision-making in a multi-agent, uncertainty-aware IPP system.
To answer this question, we compare EDL, MC Dropout, Deep Ensembles and a GP in closed loop with five planners on physics-based oil spill simulations.

\subsection{Contributions}
The contributions of this paper are fourfold. First, we show that, for deep information models, the model rather than the planner determines the outcome of uncertainty-aware multi-ASV IPP. Across three deep models, five planners and 60 physics-based oil spill simulations, the choice of model explains 80\% of the variance of the final reconstruction error against 1\% for the planner (69\% against 2\% when the GP is included). This shifts the emphasis from planner design, the focus of most IPP research, to the information model, which is usually taken as given.

Second, we introduce Evidential Deep Learning (EDL) as an information model for multi-vehicle IPP, to the best of our knowledge its first application to a system where sampling is constrained by motion, budget and multi-agent coordination. EDL delivers the lowest reconstruction error and the highest spill IoU under every planner tested, is the best-calibrated model in both the magnitude and the ranking of its uncertainty, and runs at a median of 3.9 ms per step, about 700 times faster than the GP, 35 times faster than MC Dropout and 4 times faster than Deep Ensembles. It therefore meets the accuracy, calibration and latency requirements of online replanning at once.

Third, we identify closed-loop distribution shift as a failure mode of uncertainty-aware IPP that offline evaluation does not reveal. Uncertainty-guided fleets place 49–82\% of their observations on the spill, against a median of 22\% in the training data, and under this concentration the error of Deep Ensembles grows with the budget while their uncertainty decreases. A mask-transplant experiment isolates concentration as the cause: applying the same observation masks to unrelated spills reduces the excess error of the ensemble over EDL by 27–45\%. Robustness to the data its own planner collects is therefore a requirement for an information model in closed-loop IPP.

Fourth, we separate guidance from reconstruction by reconstructing every fleet's observations with the same model, which shows when the quality of the guiding uncertainty matters. With look-ahead planners, the three deep models guide equally well; with greedy planners, the data guided by MC Dropout or Deep Ensembles are reconstructed with an 11–37\% higher error than those guided by EDL, and the GP guides worse with every planner that uses its uncertainty. This suggests that a longer planning horizon can compensate for a weaker guiding uncertainty, but not for a weaker reconstruction model.

The rest of this article is organized as follows. Section~\ref{sec:related_works}
reviews information models and planners for IPP. Section~\ref{statementproblem}
formulates the multi-vehicle monitoring problem and its constraints.
Section~\ref{methodology} presents the uncertainty decomposition, the three deep
models, the GP baseline and the five planners. Section~\ref{results}
describes the experimental setup and reports the offline accuracy and
calibration of the models, their closed-loop performance with every planner,
the separation of guidance from reconstruction, the closed-loop distribution
shift and the computation time. Section~\ref{conclusion} concludes the
article and outlines future work.

\section{Related Works} \label{sec:related_works}



Most IPP research improves the planner, from exact mixed-integer formulations \cite{dutta2025ipp} to sampling-based trees that adapt online to the belief \cite{moon2026iatigris}, \cite{fazliu2026xit}, and takes the information model as given. These planners optimize the information predicted by the model, so their benefit holds only if its uncertainty locates the error, which a GP ensures only when correctly specified \cite{dutta2025ipp}, and a single uncertainty map does not separate field variability from lack of knowledge \cite{fazliu2026xit}.

Gaussian Processes (GPs) with a stationary radial basis function (RBF) kernel are the default information model in IPP~\cite{gaussianmachine}. 
In aquatic monitoring with ASVs, they have been combined with Bayesian optimization~\cite{fedegaussian} and multi-agent reinforcement learning~\cite{amendgpDRL}. However, they present three limitations for long missions in water bodies. Exact inference scales with the number of samples $D$ as $O(D^3)$, stationary and homoscedastic noise assume uniform correlation and noise, which fails near shores or spill fronts, and the posterior variance depends on where samples were taken but only weakly on what was measured, making it a poor proxy of the error~\cite{chen2022ak}. Sparse variational GPs reduce the cost through inducing points~\cite{titsias2009variational} and have been applied to single- and multi-robot IPP~\cite{ma2017informative, jakkala2024multi}; local GPs let the correlation vary across regions~\cite{NIPS2008_01161aaa}; and non-stationary kernels~\cite{NIPS2003_326a8c05}, notably the Attentive Kernel~\cite{chen2022ak}, aim to make the variance track the prediction error. Each extension, however, targets one limitation at the cost of extra hyperparameters and per-step optimization, and the model must still be refitted online as data arrive.

Learning models have also been used for IPP ~\cite{POPOVIC2024104727}. Most learning-based works learn the policy~\cite{amendgpDRL} while keeping a GP-type belief. Learned models reconstruct a field from sparse sensors at a constant inference cost~\cite{fukami2021global, vaepoctewe}.
However, they are trained to minimize the reconstruction error and return a point estimate without a per-location uncertainty. 
In deep regression, uncertainty decomposes into aleatoric and epistemic terms~\cite{kendall2017uncertainties}. The former is learned with a heteroscedastic likelihood, while the latter is approximated by sampling, via MC Dropout~\cite{pmlrv48gal16} or Deep Ensembles~\cite{NIPS2017_9ef2ed4b}, whose cost grows with the number of passes and whose calibration degrades under distribution shift~\cite{ovadia2019can} as is the case in IPP. EDL instead obtains it in closed form by predicting the parameters of a Normal-Inverse-Gamma distribution in a single forward pass~\cite{NEURIPS2020_aab08546}, and has been applied to real-time perception and pool-based active learning~\cite{soleimany2021evidential}, although its epistemic term is a heuristic rather than an exact posterior variance~\cite{meinert2023unreasonable}.
For IPP the approximate epistemic uncertainty may still provide a useful planning signal. The planner does not compare uncertainty with a physical quantity but ranks locations by their predicted uncertainty as a proxy of expected error reduction. What guidance requires is therefore that the regions of high uncertainty coincide with the regions of high reconstruction error, not that the predicted variance
equals the error variance. We evaluate both properties separately. The overlap between high-uncertainty and high-error cells measures the uncertainty as an acquisition signal~\cite{laves2020well}. 
 
To the best of our knowledge, EDL has not been used as the information model of an IPP system, where sampling is constrained by motion, budget and coordination. Finally, deep uncertainty-aware models have been coupled with single-vehicle planning~\cite{ruckin2023informative}, while multi-agent IPP has relied on GPs~\cite{kemna2017multi, jakkala2024multi}. This paper addresses these gaps by benchmarking MC Dropout, Deep Ensembles and EDL against a GP in closed loop with five planners and a fleet of four ASVs to answer the following question. 
Does an uncertainty estimator calibrated on the training distribution remain reliable in closed-loop operation across different planners and self-collected fleet observations?

\section{Statement of the Problem} \label{statementproblem}

Consider a fleet of $N$ homogeneous vehicles that navigate a bounded environment $\mathcal{X} \subset  \mathbb{R}^2$ discretized into a 2D occupancy grid
represented as a graph $G = (\mathcal{V},E)$ where each node $v\in\mathcal{V}$ corresponds to a free cell and edges connect adjacent traversable cells. Let $\mathcal{P} = \{ p_n \mid n = 1, 2, \ldots, N \}$ be the fleet of vehicles where agent $n$ occupies a node $p_n^{(t)}\in\mathcal{V}$ at decision step $t$, and selects its next node according to a policy $\pi$.
Let $f:\mathcal{X}\rightarrow\mathbb{R}$ be an unknown scalar field describing the environmental phenomenon to be monitored (e.g., pollutant concentration or debris intensity), assumed static over the mission horizon $T$.
After $N_{\mathrm{t}}$ decision steps, the fleet has collected $\mathcal{D}=\{(x_i,y_i)\}_{i=1}^{N\cdot N_{\mathrm{t}}}$ measurements, where $x_i \in \mathcal{V}$ denotes the measurement location and $y_i \in \mathbb{R}$ is the corresponding observation according to 

\begin{equation} \label{eq:observation}
    y_i = g(f, x_i) + \varepsilon_i, \qquad \varepsilon_i \sim \mathcal{N}(0, \sigma_n^2),
\end{equation}
where $g$ is an observation model which encodes the sensor measurement process (e.g., pointwise, conic-shaped)
and $\varepsilon_i$ is zero-mean Gaussian sensor noise with variance $\sigma_n^2$,
independent across measurements and of $f$ and $x_i$.

The mission objective is to build an information model $\hat{f}$ of the environment that reconstructs the underlying environmental state $f$ from the sparse, partial observations gathered by the vehicles along their paths using a surrogate model $m(\mathcal{D})$. 
Performance is evaluated through the RMSE, Eq. (\ref{eq:metrics}) at the end of the mission, i.e., at step $T$:

\begin{equation}\label{eq:obj1}
\min \; \mathrm{RMSE}\big(f,\, \hat{f}\big) .
\end{equation}

The optimization variable is the set of paths, and IPP must provide the paths whose measurements yield the most accurate reconstruction of the field.






\subsection{Constraints}

To guarantee that missions are feasible and prevent collision configurations, the following hard constraints have been considered. 
First, node exclusivity. Only one agent can be assigned to a node in the same decision step to avoid overlaps
\begin{equation} \label{eq:constraint1}
p_m^t \neq p_n^t \quad \forall m,n \in \{1,2,\dots,N\},\; m \neq n.
\end{equation}

Second, each agent has limited battery budget. This battery is modeled as an upper limit $B$ on the total distance it can travel over the mission horizon

\begin{equation} \label{eq:constraint2}
\sum_{t=1}^{T} d\!\left(p_n^{(t-1)},p_n^{(t)}\right) \leq B
\quad \forall\, n \in \{1,\dots,N\},
\end{equation}

where $d$ is the Euclidean distance between cells.

\subsection{Assumptions}

The formulation relies on five assumptions. 
First, the initial positions of the vehicles are known and all agents depart from navigable cells in $\mathcal{V}$. 
Second, the occupancy map is known before the monitoring mission starts. Only the scalar field $f$ is unknown.
Third, the scalar field $f$ is static over the mission duration.
Fourth, localization is exact, and communication between agents is instantaneous and lossless, so all agents share a common model $\hat{f}^{(t)}$.
Fifth, measurements start only when all vehicles have reached their target positions. This excludes the transit phase, so results reflect the fleet with all vehicles in full operation.

\section{Methodology} \label{methodology}

Our approach, rather than seeking paths that explicitly minimize the reconstruction error of $\hat{f}$, adopts the predicted uncertainty $\sigma_{total}$ as a proxy of that error. This predicted uncertainty is provided by each model alongside its prediction $\hat{f}$.

We consider three deep learning models capable of quantifying
predictive uncertainty: MC Dropout \cite{pmlrv48gal16}, Deep Ensembles \cite{NIPS2017_9ef2ed4b}, and
EDL \cite{NEURIPS2020_aab08546}, and compare them against a baseline GP with an RBF kernel \cite{gaussianmachine}. These models are combined
with five planning policies: Value-Greedy, Uncertainty-Greedy, $\epsilon$-Greedy,
Receding Horizon (RH) Orienteering, and Monte Carlo Tree Search (MCTS). 



\subsection{Uncertainty Decomposition}

The information gathered by vehicles is not perfect. Onboard sensors are subject to measurement noise which introduces uncertainty into every observation, and the regions that have not been visited are only known through inference.

Thus, the uncertainty associated with $\hat{f}$ arises from two independent sources. 
Aleatoric uncertainty, $\sigma^2_{\mathrm{ale}}$, reflects the noise inherent to the physical sensor and to unmodeled sub-grid variability. 
Epistemic uncertainty $\sigma^2_{\mathrm{epi}}$ reflects the model's lack of knowledge due to limited observations. 
The predictive variance decomposes additively as
\begin{equation}\label{eq:decomposition}
    \sigma^2_{\mathrm{total}}(x) =
    \underbrace{\sigma^2_{\mathrm{ale}}(x)}_{\text{irreducible}} +
    \underbrace{\sigma^2_{\mathrm{epi}}(x)}_{\text{reducible}} .
\end{equation}


\subsection{Models for Scalar Field Reconstruction}

\begin{figure}
    \centering
    \includegraphics[width=1\linewidth]{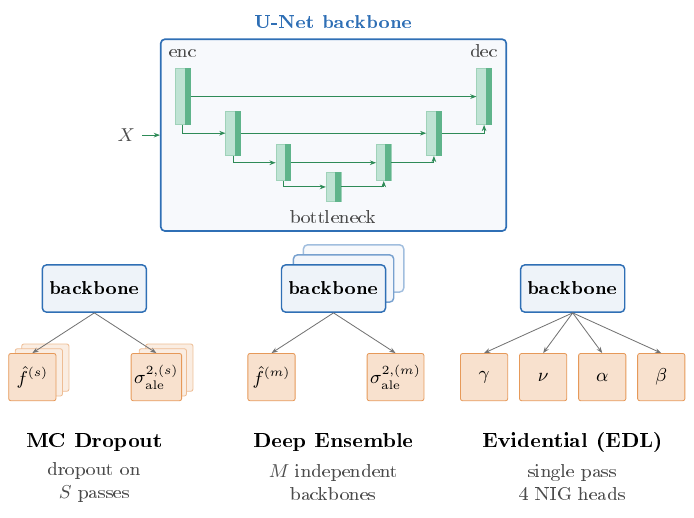}
    \caption{Uncertainty estimation methods compared in this work: MC Dropout, Deep Ensembles and EDL. All three share the same U-Net backbone (top); they differ only in how the backbone is used and in the output heads attached to it.}
    \label{fig:unet-shape}
\end{figure}

Since the map is a regular grid, the measurements can be encoded as a two-channel image $X = [U,V]\in \mathbb{R}^{2\times H \times W}$ where $U \in \{0,1\}^{H\times W}$ indicates which cells have been visited, and $V \in \mathbb{R}^{H\times W}$ is a value matrix holding the corresponding sensor readings at visited cells and zero elsewhere. Reconstruction becomes a completion from sparse pixels to a mean estimate $\hat{f}^{(m)}(x)$ and a log-variance log $\hat{\sigma}^{2(m)}_{ale}(x)$. A U-Net \cite{unet} suits the task of reconstructing an oil spill. It learns typical spill shapes from simulation, infers their extent from a few transects, and preserves boundaries through skip connections, at a cost independent of the number of samples. A stationary GP instead assumes a single correlation length and smooths the boundaries. MC Dropout, Deep Ensembles and EDL share the same U-Net, so they differ only in how they estimate uncertainty. Figure \ref{fig:unet-shape} summarizes the shared backbone and the model-specific heads.


Calibration training is carried out using the heteroscedastic negative log likelihood defined in Eq. (\ref{eq:negativeLL}), which incentivizes accurate reconstruction and calibrated predictive variance. 
The log term penalizes overestimating $\sigma^2$, while the squared residual avoids underestimating. Obstacles and shorelines are sorted out via a navigable waters mask $\mathcal{X}_{nav}$. The loss is averaged over the navigable cells

\begin{equation} \label{eq:negativeLL}
\mathcal{L}_{\text{NLL}} = \frac{1}{|\mathcal{X}_{nav}|} \sum_{x \in \mathcal{X}_{nav}}
\left[ \frac{\big(f(x) - \hat{f}(x)\big)^2}{2\hat{\sigma}^2_{ale}(x)}
+ \frac{1}{2}\log \hat{\sigma}^2_{ale}(x) \right],
\end{equation}


\subsubsection{Monte Carlo Dropout}
MC Dropout\cite{pmlrv48gal16} inserts dropout layers with rate $p$ after each convolutional block. Dropout remains active at inference, and $S$ stochastic forward passes yield a set of predictions $\{ \hat{f}^{(s)}\}^S_{s=1}$ interpreted as samples from an approximate posterior over network weights, from which epistemic uncertainty and the predictive mean are estimated as:  
\begin{equation}
\hat{f}(x) = \frac{1}{S}\sum_{s=1}^{S} \hat{f}^{(s)}(x) \qquad
\hat{\sigma}^2_{\mathrm{ale}}(x) = \frac{1}{S}\sum_{s=1}^{S} \hat{\sigma}^{2(s)}_{\mathrm{ale}}(x)
\end{equation}
\begin{equation}
\sigma^2_{\mathrm{epi}}(x) = \frac{1}{S}\sum_{s=1}^{S} \left(\hat{f}^{(s)}(x) - \hat{f}(x)\right)^2.
\end{equation}


\subsubsection{Deep Ensembles}
Deep Ensembles \cite{NIPS2017_9ef2ed4b} train $M$ independent instances of the backbone, each with a different random initialization of their parameters and a different shuffling of the training data. Each instance includes an auxiliary log-variance head.

\begin{equation}
    \hat{f}(x) = \frac{1}{M}\sum_{m=1}^{M} \hat{f}^{(m)}(x) \quad
    \hat{\sigma}^2_{\mathrm{ale}}(x) = \frac{1}{M}\sum_{m=1}^{M} \hat{\sigma}^{2(m)}_{\mathrm{ale}}(x)
\end{equation}
\begin{equation}
    \sigma^2_{\mathrm{epi}}(x) = \frac{1}{M}\sum_{m=1}^{M}\left(\hat{f}^{(m)}(x) - \hat{f}(x)\right)^2.
\end{equation}


\subsubsection{Evidential Deep Learning}
EDL \cite{NEURIPS2020_aab08546} places a Normal-Inverse-Gamma prior over the Gaussian likelihood, modifying the backbone to output four quantities $(\gamma, \nu, \alpha, \beta)$ per spatial location, with constraints $\nu > 0, \ \alpha > 1, \ \beta > 0$ enforced through softplus activations. Both uncertainty components are available in closed form:
\begin{equation} \label{eq:edl_params}
    \hat{f}(x) = \gamma,\qquad
\sigma^2_{\mathrm{epi}}(x) = \frac{\beta}{\nu(\alpha - 1)},\qquad
\sigma^2_{\mathrm{ale}}(x) = \frac{\beta}{\alpha - 1}.
\end{equation}
Training minimizes the negative log-likelihood of the ground truth under the Student-$t$ distribution $\mathrm{St}\!\left(\gamma, \frac{\beta}{\nu\alpha}, 2\alpha\right)$ obtained by marginalizing the NIG prior over the variance
\begin{equation}
\begin{split}
    \mathcal{L}_{\mathrm{NIG}} ={}& \frac{1}{2}\log\frac{\pi}{\nu} - \alpha\log\Omega  \left(\alpha + \tfrac{1}{2}\right)\log\!\left(\nu\,(f - \gamma)^2 + \Omega\right) \\
    &+ \log\frac{\Gamma(\alpha)}{\Gamma\!\left(\alpha + \tfrac{1}{2}\right)},
\end{split}
\label{eq:nigll}
\end{equation}

with $\Omega = 2\beta(1 + \nu)$ and $f$ the ground-truth field, plus an evidence regularizer $\lambda$ that penalizes high evidence on incorrect predictions
\begin{multline}
    \mathcal{L}_{\mathrm{EDL}} = \frac{1}{|\mathcal{X}_{nav}|}\sum_{x \in \mathcal{X}_{nav}} \Big[ \mathcal{L}_{\mathrm{NIG}}(x) \\
    + \lambda \left| f(x) - \gamma(x) \right| \big( 2\nu(x) + \alpha(x) \big) \Big].
    \label{eq:edl_loss}
\end{multline}

\subsubsection{Gaussian Process Baseline}
As a probabilistic baseline, we employ a GP \cite{gaussianmachine} with composite kernel
\begin{equation} \label{eq:gp_kernel}
    k(x, x') = \sigma_f^2 \exp\!\left(-\frac{\|x - x'\|^2}{2\ell^2}\right) + \sigma_n^2\,\delta(x, x'),
\end{equation}

where the first term models the spatial correlation of the field and the second the homoscedastic measurement noise.
At each time step $t$, the hyperparameters $\{\sigma_f^2, \ell, \sigma_n^2\}$ are optimized by maximizing the marginal log-likelihood (Type II) as in Eq. (\ref{eq:gp_mll}) of the sparse samples collected so far using an L-BFGS optimizer. Observations are standardized as $\tilde{\mathbf{y}}=(y-\bar{y})/s$
where $\bar{y}$ and $s$ are the empirical mean and standard deviation of the measurements in $\mathcal{D}$.

\begin{align}
\log p(\tilde{\mathbf{y}} \mid X, \boldsymbol{\theta}) &= -\tfrac{1}{2}\tilde{\mathbf{y}}^\top K^{-1}\tilde{\mathbf{y}} - \tfrac{1}{2}\log|K| - \tfrac{N_t\cdot N}{2}\log 2\pi, \label{eq:gp_mll}\\
\hat{f}(x) &= \bar{y} + s\, \mathbf{k}(x)^\top K^{-1}\tilde{\mathbf{y}}, \label{eq:gp_mean}\\
\sigma^2_{\mathrm{epi}}(x) &= s^2 \left(k(x,x) - \mathbf{k}(x)^\top K^{-1}\mathbf{k}(x)\right), \label{eq:gp_epi}\\
\sigma^2_{\mathrm{ale}} &= s^2\, \sigma_n^2. \label{eq:gp_ale}
\end{align}

where $[K]_{ij} = k(x_i, x_j)$ is the Gram matrix over the observed locations, with the noise term contributing only to its diagonal, $k(x, x) = \sigma_f^2 + \sigma_n^2$, and $\mathbf{k}(x) = [k(x, x_1), \dots, k(x, x_{N_t})]^\top$ is the cross-covariance between a query point and the observations, from which the noise term is excluded. 
The posterior variance defined in Eq. \eqref{eq:gp_epi} is used as epistemic uncertainty: it contracts at sampled locations, where it is bounded below by the noise floor $s^2\sigma_n^2$, and recovers the prior variance $s^2(\sigma_f^2 + \sigma_n^2)$ far from them. The noise variance defined in Eq. \eqref{eq:gp_ale} provides a spatially uniform aleatoric estimate.











\subsection{Path Planning Algorithms}

All planners described in this paper operate inside a common receding-horizon loop planning strategy in which, at decision step $t$, a trajectory is optimized over a finite look-ahead horizon $T_H$, but only the first action is executed before the problem is solved again with updated information. At decision step $t$, the model $m(\mathcal{D}_t)$ is evaluated obtaining $\hat{f}_t(x)$ and $\hat{\sigma}^2_{t}(x)$. The planner uses this information, together with the remaining budget and proposes a candidate trajectory for the next $T_H$ steps. Then, the vehicles move, measure, the data is updated to $\mathcal{D}_{t+1}$, and a new receding-horizon loop starts. As a result, every decision is based on the most recent observations, and the executed path adapts continuously instead of committing to a plan computed from the initial state. All planners described in this paper receive identical $X$ inputs and have the same re-planning frequency, they differ only in how they compute the proposed trajectory, which allows a direct comparison between optimization strategies. The uncertainty map of the unobserved cells is normalized to [0, 1] by min-max scaling. The planners are therefore invariant to the scale of the uncertainty and respond only to where it is high relative to the rest of the map.

\begin{figure*}
    \centering
    \includegraphics[width=\linewidth]{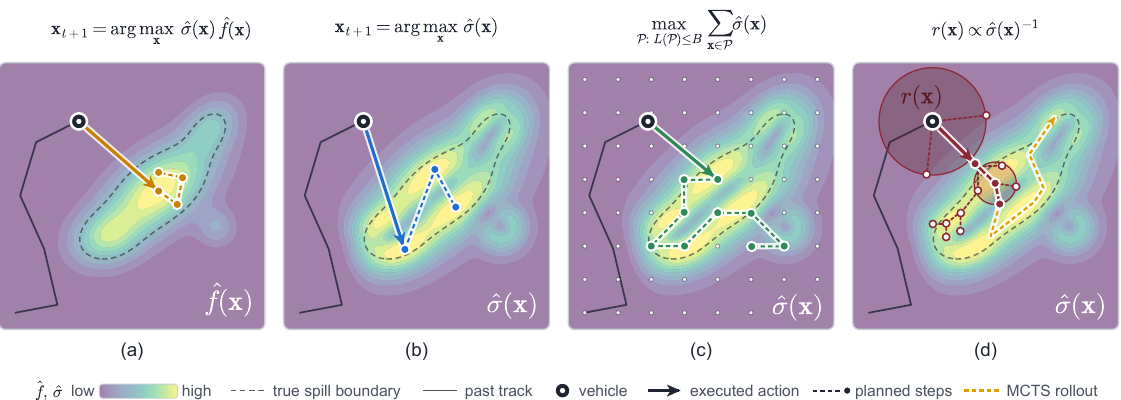}
    \caption{Summary of the four planning strategies discussed: (a) Value-Greedy planner, (b) Uncertainty-Greedy planner, (c) RH Orienteering, and (d) MCTS with uncertainty-adaptive branching.}
    \label{fig:strategies_overview}
\end{figure*}

The planning problem is defined over the joint action space of the fleet.
Optimizing the joint action $\mathbf{a} = (a_1, \dots, a_N)$ is infeasible, as the joint action space grows $\mathcal{O}(|\mathcal{A}_n|^N)$, while optimizing each $a_n$ independently ignores the coupling between vehicles and leads to redundant measurements \cite{claus1998dynamics}.
Following the principle of prioritized sequential planning
\cite{prioseqplanning2015}, a simulated precedence mechanism is adopted: vehicles decide in a fixed priority order, and vehicle $n$ best-responds to the actions already committed by its predecessors,
\begin{equation} \label{eq:best_response}
    a_n^* = \arg\max_{a_n \in \mathcal{A}_n}
    J_n\!\left(a_n \mid \mathcal{D}, U^{(n)}\right),
    U^{(n)} = U \cup \{a_1^*, \dots, a_{n-1}^*\},
\end{equation}
where $J_n$ is the utility of the planner and $U^{(n)}$ treats the targets of the predecessors as already visited. The cost thus reduces to $\mathcal{O}(N|\mathcal{A}_n|)$, and the fleet
spreads without any explicit inter-vehicle distance constraint. In Fig. \ref{fig:strategies_overview}, a visual summary of the following planners can be seen.

\subsubsection{Greedy planners}

Three greedy variant policies are considered in this paper. Greedy planners select the next waypoint $p_n^{t+1}$ of each vehicle by maximizing a scalar acquisition function over the set of reachable, unvisited cells, without look-ahead
\begin{equation}
\mathcal{C}_n^t=\big\{x\in\mathcal{X}:\ \lVert x-p_n^t\rVert_2\le r,\ x\notin\mathcal{N}(\mathcal{X}_k),\ x\neq w_j^{t+1}\ \forall j<n\big\},
\label{eq:candidates}
\end{equation}
where $\mathcal{X}_k$ is the set of visited cells, $\mathcal{N}(\cdot)$ adds their 8-neighbors, and $r=30$ cells. The next position is selected with

\begin{equation}
    p_n^{t+1} = \arg\max_{x \in \mathcal{C}_n^t} a(x),
\end{equation}

where $a(x)$ is the acquisition function.  The Uncertainty-Greedy planner sets $a(x)=\sigma_{total}(x)$ leading vehicles towards the least known location, aiming to reduce uncertainty. The Value-Greedy planner sets $a(x) = \hat{f}(x)\cdot\sigma_{total}(x)$, leading the vehicles towards areas with predicted field value and high uncertainty, serving an exploitation objective that tracks the core of the phenomenon. The $\epsilon$-Greedy planner follows $a(x) = \hat{f}(x)$ with probability $\epsilon$ and takes random actions with a probability $1-\epsilon$, balancing exploitation of the spill and exploration. 


The greedy variants are myopic but computationally negligible, and they remain a strong baseline because each measurement reduces the uncertainty left for subsequent ones.

\subsubsection{Monte Carlo tree search}
MCTS \cite{MCTS} replaces the myopic greedy criterion with a multi-step look-ahead built through a stochastic tree expansion. At each
planning invocation, a search tree is rooted at the current state $(x_t)$ and the four standard phases are iterated: selection via the upper confidence bounds criterion \cite{kocsis2006bandit}, expansion of an unvisited child, rollout under the Uncertainty-Greedy heuristic, and backpropagation of the cumulative uncertainty collected along the simulated path.
After a fixed budget of $I_{\mathrm{MCTS}}$ iterations, the fleet commits to the first action of the highest value branch, consistent with the receding-horizon loop.

To avoid the combinatorial growth of the action space, the fleet decisions are sequentialized. Each tree level corresponds to the action of a single agent, so a joint fleet action spans $N$ consecutive levels. The order in which agents choose is set by the local uncertainty around each vehicle, thus making the most informative decisions near the root. 

Additionally, an uncertainty-adaptive branching strategy has been taken, where the neighborhood radius $r(x)$ from which the children of a node at location $x$ are sampled is made inversely proportional to the local uncertainty

\begin{equation}
r(\mathbf{x}) = \min\left(\frac{r_{\max} - r_{\min}}{\hat{\sigma}_{\text{total}}(\mathbf{x}) + \epsilon_r} + r_{\min}, r_{\max}\right).
\end{equation}
where $\epsilon_r > 0$ ensures numerical stability and $r_{\min}$, $r_{\max}$  are predefined bounds. 
Search resolution is thus allocated by the same signal that defines the objective without enlarging the branching factor.

\subsubsection{Receding Horizon Orienteering}

RH Orienteering formulation treats planning as a combinatorial optimization problem \cite{vansteenwegen2011orienteering}. Candidate waypoints are sampled on a regular sub-grid of the map, and the planner selects an ordered subset that maximizes total reward while respecting the budget constraints on trajectory cost. The reward of traversing an edge is defined as the sum of uncertainty values over the pixels of the connecting segment, and the cost as the Euclidean distance between waypoints. Defining reward along edges rather than at cells credits the information gathered in transit, and lets the planner reason about the collective value of a sequence of waypoints instead of committing one step at a time.

The problem is solved with a two-phase metaheuristic. First, a multi-start greedy construction generates a diverse set of initial paths according to reward-to-cost efficiency under random orderings of the candidate list. Second, a hill-climbing local search refines the best solution by alternating between removal of interior waypoints, which frees the budget, and insertion of unvisited candidates whenever they increase total reward within the budget, until no single move improves it.

\section{Experimental Results} \label{results}

We design the experiments to answer the question of Section~\ref{sec:related_works} in four steps. First, we measure the accuracy and calibration of each model on held-out spills observed as in training (Sec.~\ref{sec:res_offline}). Second, we close the loop: we cross the four models with the five planners and measure the calibration, the reconstruction error and the spill IoU at the end of each mission (Secs.~\ref{sec:res_calib}-\ref{sec:res_planning}). Third, we separate the two roles of the model. We follow the error along the mission and under smaller budgets (Sec.~\ref{sec:res_budget}), and we reconstruct every fleet's observations with every model, which isolates the quality of the samples from that of the reconstructor (Sec.~\ref{sec:res_guidance}). Fourth, we test robustness: we compare each model on its own fleet's data with the same samples placed on unrelated spills (Sec.~\ref{sec:res_shift}). Finally, we study the real-time constraints of every model and path planner subjected to a real fleet movement capacity (Sec.~\ref{sec:res_cost}).


\subsection{Experimental setup}

We simulate 2000 spills on a $100\times100$ grid. Ground truths come from a Lagrangian particle model where particles are released at a random source cell and displaced by turbulent diffusion, a uniform wind drift and a divergence-free tidal current. The ground truth of each cell is normalized to $[0,1]$, where 0 is clean water and 1 a cell fully covered by oil. Each training sample observes the spill with a random planner, creating a dataset split of 1600 for training and 400 for testing. Later on, we evaluate the 20 model-planner pairs online on the same 60 spills with a budget of 300 cells of travel per vehicle, and 50, 100 and 150  on 30 further spills. Training and evaluation splits differ on seed and spill conditions.
The measurements gathered by vehicles present noise following a normal distribution $\varepsilon_i \sim \mathcal{N}(0, 0.01^2)$.

We tuned the three deep models with Optuna~\cite{akiba2019optuna} under the same protocol and architecture (50 epochs with batch size 16 and Adam). The objective is the mean negative log-likelihood on the unobserved cells of 320 validation spills held out from the 1600 training spills, using the likelihood each model is trained with: the Gaussian NLL with $\sigma_{\mathrm{total}}$ for MC Dropout ($p=0.2$ and $S=30$) and Deep Ensembles ($M=5$), and the Normal-Inverse-Gamma NLL for EDL (with $\lambda=0$ at a first hyperparameter optimization, and a second round for $\lambda = 1.1\cdot10^{-4}$). Every method received the same budget of 50 trials (Tree-structured Parzen Estimator sampler with median pruning). The budget is equal in trials but not in compute, because each ensemble trial trains all its members (1.9 GPU-hours against 0.4-0.7 for the other methods). The GP has no tuning stage: it fits its kernel hyperparameters by maximum marginal likelihood at every step.

The greedy planners choose among the cells within a radius of 30; $\varepsilon$ decays from 0.1 to 0.01 over 100 steps; RH Orienteering plans paths of up to 30 cells; and MCTS uses a depth of 50 with 500 iterations. At the end of each mission we measure the RMSE on the set $\mathcal{X}_u$ of unobserved cells and the IoU between the true and predicted spill masks, both thresholded at $\tau=0.05$:
\begin{equation}
\mathrm{RMSE}=\sqrt{\frac{1}{|\mathcal{X}_u|}\sum_{x\in\mathcal{X}_u}\big(f(x)-\hat f(x)\big)^2},\qquad
\mathrm{IoU}=\frac{|\hat S\cap S|}{|\hat S\cup S|}.
\label{eq:metrics}
\end{equation}
We run the experiments on a server with Ubuntu 22.04.4 LTS, two Intel Xeon Gold 5220R CPUs at 2.20\,GHz, 192\,GB of RAM, and two GPUs (Nvidia RTX A6000 48\,GB and RTX 3090 24\,GB). 

\subsection{Offline results}\label{sec:res_offline}

Before closing the loop, we evaluate each model on the 400 held-out spills, observed as in training along random paths (ranging 1.2-5\% coverage). Every model improves with more visited cells. EDL is the most accurate and the best calibrated model, and GP the least accurate, with the largest calibration deviation (Table~\ref{tab:offline}; the deviation is defined in Sec.~\ref{sec:res_calib}). Between them, Deep Ensembles are more accurate than MC Dropout (RMSE 0.084 against 0.100) and similarly calibrated. Deep Ensembles also become better calibrated with more samples. 


\begin{table}[t]
\centering
\footnotesize
\caption{Offline results on the 400 test spills (mean over spills), observed along random paths as in training. All metrics on the unobserved navigable cells with $\sigma=\sigma_{\mathrm{total}}$. Cal.: mean deviation of the calibration curve from the diagonal (Sec.~\ref{sec:res_calib}). UCE scaled by $10^{3}$. Best result in bold.}
\label{tab:offline}
\setlength{\tabcolsep}{3pt}
\begin{tabular}{llcccccc}
\toprule
 Model & RMSE$\downarrow$ & IoU$\uparrow$ & NLL$\downarrow$ & UCE$\downarrow$ & Cal.$\downarrow$ & $\bar\sigma$ \\
\midrule
 EDL       & \textbf{0.036} & \textbf{0.88} & \textbf{$-$3.16} & \textbf{1.85} & \textbf{0.011} & 0.027 \\
 Deep Ensembles & 0.084 & 0.61 & $-$2.53 & 8.13  & 0.084 & 0.080 \\
 MC Dropout  & 0.100 & 0.57 & $-$2.08 & 11.56 & 0.081 & 0.101 \\
 GP        & 0.118    & 0.38   & 8.41      & 11.33    & 0.129    & 0.074    \\ 
\bottomrule
\end{tabular}
\end{table}



\subsection{Uncertainty calibration}\label{sec:res_calib}

\begin{figure}[t]
\centering
\includegraphics[width=\columnwidth]{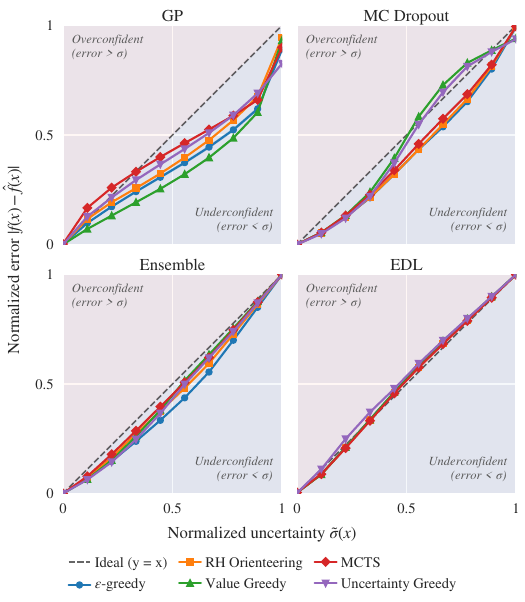}
\caption{Calibration of the uncertainty estimates per model, one line per planner. For each prediction, $\sigma_{\mathrm{total}}$ and the absolute error $|f(x)-\hat{f}(x)|$ on the unobserved cells are min-max normalized; cells are grouped into deciles of $\tilde{\sigma}$, and each point is the mean normalized error of a decile. Points on the diagonal indicate perfect calibration; above it the model is overconfident, below it underconfident.}
\label{fig:calib}
\end{figure}

We characterize the uncertainty of each model with two measures. The uncertainty calibration error (UCE)~\cite{laves2019UCE} compares the magnitude of the predicted variance with the squared error. The cells of $\mathcal{X}_u$ are grouped into $N_b=10$ equal-width bins $U_b$ of $\sigma^2$, and
\begin{equation}
\mathrm{UCE}=\sum_{b=1}^{N_b}\frac{|U_b|}{|\mathcal{X}_u|}\,\big|\,\overline{\sigma^2}_b-\overline{e^2}_b\big|,
\label{eq:uce}
\end{equation}
where $\overline{\sigma^2}_b$ and $\overline{e^2}_b$ are the mean predicted variance and the mean squared error $(f(x)-\hat f(x))^2$ of the cells in $U_b$. We compute it at the end of each mission with $\sigma=\sigma_{\mathrm{total}}$. The calibration curves of Fig.~\ref{fig:calib} instead normalize $\sigma$ and $|f(x)-\hat f(x)|$ per prediction,
group the cells into deciles of $\tilde\sigma$, and plot the mean normalized error of each decile. Their mean distance to the diagonal (calibration deviation) measures whether $\sigma$ ranks the error, regardless of its scale.

The two measures answer different questions. EDL is the best calibrated model on both, under every planner: its UCE ($0.7$-$0.9\cdot10^{-3}$, Table~\ref{tab:uce}) is 2.4 to 100 times lower than that of the other models, and its curves deviate from the diagonal by 0.011-0.020. The Deep Ensembles have the highest UCE under every planner ($15.9$-$71.5\cdot10^{-3}$, highest with MCTS), although their curves stay close to the diagonal (0.031-0.067): their $\sigma$ ranks the error, but its magnitude does not match it. The GP shows the opposite pattern. Its UCE is moderate ($1.9$-$7.2\cdot10^{-3}$), because its large $\sigma$ matches its large error on average, but its curves deviate most under every planner (0.089-0.161), so its $\sigma$ does not locate the error. MC Dropout is intermediate on both (UCE $9.3$-$12.4\cdot10^{-3}$, deviation 0.049-0.087).

Regarding EDL, Eq. (\ref{eq:edl_params}) implies that both uncertainty terms differ by a per-location factor, $\sigma^2_{\mathrm{epi}}=\sigma^2_{\mathrm{ale}}/\nu$, which questions the reliability of $\sigma_{\mathrm{epi}}$ as a posterior variance. However, this does not prevent the EDL uncertainty from being a useful acquisition signal.


\begin{table}[t]
\centering
\footnotesize
\caption{UCE per planner and model at the end of the mission, reported as mean (std) over 60 maps. All values are scaled by $10^{3}$; lower is better. Best result per planner in bold.}
\label{tab:uce}
\setlength{\tabcolsep}{3pt}
\begin{tabular}{lcccc}
\toprule
Planner & GP & MC Drop. & Deep Ens. & EDL \\
\midrule
Value Greedy         & 7.2 (6.3)  & 9.3 (4.2)  & 24.0 (14.8) & \textbf{0.7 (0.3)} \\
Uncertainty-Greedy   & 4.0 (4.1)  & 11.4 (3.7) & 23.3 (15.6) & \textbf{0.8 (0.7)} \\
$\varepsilon$-Greedy & 5.6 (4.8)  & 12.4 (3.8) & 15.9 (11.9) & \textbf{0.9 (0.2)} \\
RH Orienteering      & 1.9 (2.5)  & 10.7 (4.1) & 31.6 (19.7) & \textbf{0.8 (0.3)} \\
MCTS                 & 5.3 (16.2) & 11.5 (4.0) & 71.5 (48.4) & \textbf{0.7 (0.2)} \\
\bottomrule
\end{tabular}
\end{table}

\subsection{Reconstruction and planning}\label{sec:res_planning}

\begin{figure}[t]
\centering
\includegraphics[width=\columnwidth]{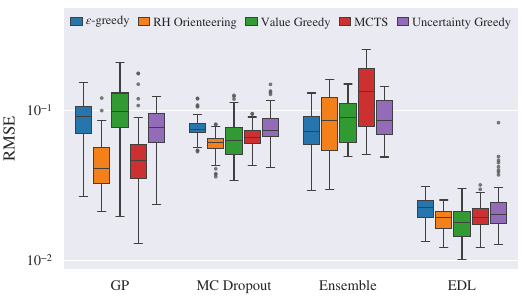}
\caption{RMSE on the unobserved navigable cells at the end of each episode, per model and planner (log scale); 60 maps per model and planner.}
\label{fig:rmse}
\end{figure}

\begin{figure}[t]
\centering
\includegraphics[width=\columnwidth]{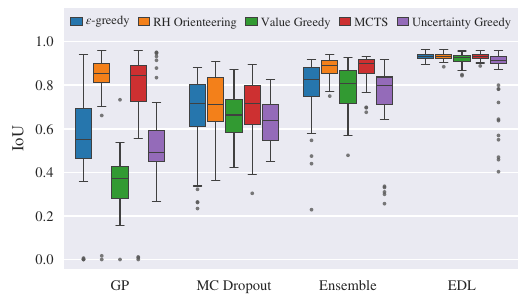}
\caption{IoU between the reconstructed and the true spill regions ($\tau=0.05$) over the navigable cells at the end of each episode, per model and planner; 60 maps per model and planner.}
\label{fig:iou}
\end{figure}

For the deep models, the reconstruction model, not the planner, determines the outcome (Figs.~\ref{fig:rmse}-\ref{fig:iou}). Across the three deep models, five planners and 60 maps, the model explains 80.2\% of the variance of the log final RMSE and 45.4\% of that of the IoU; the planner explains 1.1\% and 5.4\%. With EDL the five planners end between 0.018 and 0.023 of median RMSE, while for a fixed planner the median RMSE spans 0.018 to 0.13 across the deep models. No planner is best in RMSE for every model: Value-Greedy is best with EDL, RH Orienteering with MC Dropout and the GP, and $\varepsilon$-Greedy with the Deep Ensembles, for which MCTS is the worst.

The GP depends much more on the planner. With RH Orienteering and MCTS it reaches a median RMSE of 0.041 and 0.047 and an IoU of 0.85 and 0.84, more accurate than MC Dropout and the Deep Ensembles with any planner, though still twice the error of EDL. With the greedy planners and $\varepsilon$-Greedy its median RMSE is 0.077-0.098. Including the GP, the model explains 69.0\% of the variance of the log RMSE, the planner 1.9\% and their interaction 6.0\%.

In IoU, MCTS and RH Orienteering delineate the spill better than both greedy planners with every model.
The one exception is MCTS against Value-Greedy with MC Dropout. 
The two look-ahead planners do not differ significantly with any model. $\varepsilon$-Greedy matches them with EDL and MC Dropout and falls below them with the Deep Ensembles and the GP. MCTS, RH Orienteering and $\varepsilon$-Greedy also observe more cells (840-1030 against 640-860 for the greedy planners), which suggests that coverage drives spill delineation.

These end-to-end comparisons change two things at once: the paths followed by the fleet and the model that reconstructs the field from the samples. The next three subsections separate both effects and explain the failure of the Deep Ensembles.

\subsection{Mission evolution and budget}\label{sec:res_budget}

\begin{figure}[t]
\centering
\includegraphics[width=\columnwidth]{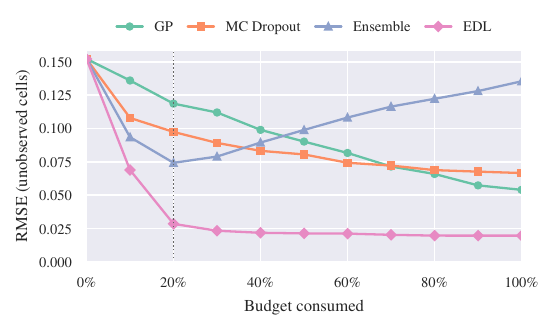}
\caption{RMSE on the unobserved navigable cells along the mission, MCTS with budget 300, mean over 60 maps; each model reconstructs the observations of its own fleet. Dotted line: 20\% of the budget.}
\label{fig:evolution}
\end{figure}

EDL reaches its final accuracy early in the mission and with small budgets. We align each MCTS episode on the fraction of budget consumed by the agent that has travelled furthest (Fig.~\ref{fig:evolution}). EDL reaches an RMSE of 0.029 at 20\% of the budget, already below the lowest value that any other model reaches during the mission (GP 0.054, MC Dropout 0.067, Deep Ensembles 0.074). The error of the Deep Ensembles is lowest at 20\% of the budget and then rises to 0.135 at the end.
Reconstructed by EDL, the observations of the Deep-Ensemble fleet keep improving until the end (0.020), so the degradation lies in the ensemble's reconstruction, not in the paths. 

With 50 cells per vehicle, EDL reaches an RMSE of 0.031, lower than MC Dropout (0.081) and the Deep Ensembles (0.122) with 150 cells on the same maps (Table~\ref{tab:budget}). The error of the Deep Ensembles grows with the budget, consistently with their degradation along the mission. The GP, in contrast, keeps improving until the end. Its error falls along the mission (from 0.136 at 10\% of the budget to 0.054) and with the budget (0.123 with 50 cells, 0.085 with 150), and its IoU rises from 0.23 to 0.78. It remains limited by its reconstruction, since EDL reconstructs the same observations with an RMSE of 0.024. Its paths are also less informative (Sec.~\ref{sec:res_guidance}).

\begin{table}[t]
\centering
\footnotesize
\caption{MCTS with each model at the end of the mission, mean over maps (budgets 50-150: maps 60-89; budget 300: maps 0-59).}
\label{tab:budget}
\setlength{\tabcolsep}{2.5pt}
\begin{tabular}{lcccccccc}
\toprule
 & \multicolumn{4}{c}{RMSE} & \multicolumn{4}{c}{IoU} \\
\cmidrule(lr){2-5}\cmidrule(lr){6-9}
$B$ & EDL & MCD & DE & GP & EDL & MCD & DE & GP \\
\midrule
50  & \textbf{0.031} & 0.102 & 0.083 & 0.123 & \textbf{0.88} & 0.52 & 0.58 & 0.35 \\
100 & \textbf{0.024} & 0.086 & 0.097 & 0.110 & \textbf{0.93} & 0.58 & 0.69 & 0.41 \\
150 & \textbf{0.023} & 0.081 & 0.122 & 0.085 & \textbf{0.92} & 0.60 & 0.77 & 0.55 \\
300 & \textbf{0.020} & 0.067 & 0.135 & 0.054 & \textbf{0.93} & 0.70 & 0.88 & 0.78 \\
\bottomrule
\end{tabular}
\end{table}

\subsection{Guidance versus reconstruction}\label{sec:res_guidance}

With the same reconstruction model, the paths guided by the three deep models are comparably informative. To isolate the paths, we reconstruct the observations collected by each fleet with every model, so that differences for a fixed reconstructor come only from where the fleet sampled. With MCTS at budget 150,
EDL reconstructs the data of the EDL-, MC Dropout- and Deep-Ensemble-guided fleets with an RMSE of 0.0228, 0.0252 and 0.0237, whereas reconstructing the same EDL-guided data with MC Dropout or the Deep Ensembles gives 0.0719 and 0.1261. Changing the paths changes the RMSE by a few percent; changing the reconstruction model for the same samples changes it by a factor of 3.2 to 5.5. The same holds at budget 300,
where the three deep fleets reconstructed by EDL reach 0.0196, 0.0203 and 0.0200.

We tested, before running these experiments, whether EDL-guided paths yield a lower final RMSE than the paths guided by each other deep model when all are reconstructed by EDL.
Only the comparison with MC Dropout at budget 150 is significant (EDL-guided paths 7.0\% lower in median).
In the other five comparisons the median difference lies between $0.6\%$ and $3.7\%$.
The outcome is the same with the RMSE over the whole grid, as originally specified. The fleets visit largely different cells, with an overlap of the visited cells of 0.18-0.20 (Jaccard),
yet collect equally informative samples. Because the field is spatially smooth, many different paths are close to optimal.

The same analysis on all five planners at budget 300 shows that guidance matters only for greedy planners (Table~\ref{tab:guidance}). Reconstructed by EDL, the guiding model explains 3.9\% of the variance of the final error of the deep fleets, against 80.2\% when each model reconstructs its own data. With Uncertainty-Greedy and Value-Greedy, the data guided by the Deep Ensembles or MC Dropout are reconstructed with an 11-37\% higher median error than the EDL-guided data.
With MCTS, RH Orienteering and $\varepsilon$-Greedy, the three deep models guide equally well (differences $\le3\%$, not significant). $\varepsilon$-Greedy, which does not use $\sigma$, acts as a control. Look-ahead planners therefore absorb differences in the quality of the guiding uncertainty that greedy planners pass on to the data.

A GP guides worse with every planner that uses $\sigma$. We fixed this test before running the experiments. 
Reconstructed by EDL, the data of GP-guided fleets have a higher final RMSE than those of EDL-guided fleets with MCTS (11\% worse),
RH Orienteering (18\%),
Uncertainty-Greedy (23\%)
and Value-Greedy (11\%)
at budget 300 
and with MCTS at budgets 50, 100 and 150 (25\%, 21\%, 19\%).
GP-guided fleets observe as many cells or more, but only 24-52\% of them on the spill, against 55-83\% under EDL. The GP variance depends more significantly on where the samples are, not on their values, so it spreads the fleet over unvisited water instead of the spill. With $\varepsilon$-Greedy the gap shrinks to 5\%.

\begin{table}[t]
\centering
\footnotesize
\caption{RMSE on the unobserved navigable cells at the end of the mission when every fleet's observations are reconstructed by EDL; budget 300, mean over 60 maps.}
\label{tab:guidance}
\setlength{\tabcolsep}{3pt}
\begin{tabular}{lcccc}
\toprule
 & \multicolumn{4}{c}{Fleet guided by} \\
\cmidrule(lr){2-5}
Planner & EDL & Deep Ens. & MC Drop. & GP \\
\midrule
MCTS                 & 0.0196 & 0.0200 & 0.0203 & 0.0240 \\
RH Orienteering      & 0.0190 & 0.0192 & 0.0196 & 0.0229 \\
Uncertainty-Greedy   & 0.0235 & 0.0292 & 0.0262 & 0.0260 \\
Value-Greedy         & 0.0184 & 0.0251 & 0.0258 & 0.0220 \\
$\varepsilon$-Greedy & 0.0220 & 0.0233 & 0.0228 & 0.0237 \\
\bottomrule
\end{tabular}
\end{table}

\subsection{Closed-loop distribution shift}\label{sec:res_shift}

Uncertainty-guided sampling drives the observations away from the training distribution, and the Deep Ensembles fail under that shift. In the training set each sample has 82-570 observed cells, of which a median 22\% lie on the spill (ground truth above 0.05; 95th percentile 62\%). At the end of the mission every fleet observes 640-1030 cells. Table~\ref{tab:shift} shows that fleets guided by the deep models place 49-82\% of them on the spill, against 30-32\% with $\varepsilon$-Greedy and 24-49\% for the GP-guided fleets, which place fewer than the deep-guided fleets with every planner that uses $\sigma$.

The concentration, not the amount of data, drives the failure. Within each map and at a fixed number of observed cells, the error of the Deep Ensembles on the unobserved cells grows with the fraction of observations on the spill
and not with the number of cells.
Their error on the cells already observed grows much more,
up to 0.45-0.54 with Value-Greedy, while their mean uncertainty falls as observations accumulate.
EDL improves with more cells 
and is not harmed by the concentration. 

To isolate the concentration, we apply each fleet's observation mask, with the same paths and the same cells, to the ground truth of another map, which lowers its overlap with the spill (Table~\ref{tab:shift}). The excess error of the Deep Ensembles over EDL falls significantly (the exception is the Deep-Ensemble-guided Uncertainty-Greedy fleet).
It also falls for the GP-guided Value-Greedy fleet, the only GP-guided fleet whose overlap with the spill drops (0.49 to 0.33). It does not change for $\varepsilon$-Greedy or for the other GP-guided fleets, whose overlap barely changes. This explains why MCTS is the worst planner with the Deep Ensembles. Guided by a deep model, MCTS places about half of its observations on the spill; guided by the GP it places 24\%, as many as it would place on an unrelated spill, and the ensemble then reconstructs its data much better (RMSE 0.066, against 0.107-0.136 for the deep-guided MCTS fleets). An information model for uncertainty-aware IPP must therefore be robust to the observations its own planner collects: EDL is, and the Deep Ensembles are not.

\begin{table}[t]
\centering
\footnotesize
\caption{Mask transplant (budget 300, 60 maps): each fleet's observation mask applied to its own spill and to another map. Excess error: RMSE of the Deep Ensembles minus RMSE of EDL on the unobserved cells (mean over maps); $p$: Wilcoxon paired by map.}
\label{tab:shift}
\setlength{\tabcolsep}{1.5pt}
\begin{tabular}{lccc}
\toprule
Fleet & On spill & Excess error & $p$ \\
 & own $\to$ other & own $\to$ other & \\
\midrule
EDL-guided MCTS            & 0.54 $\to$ 0.34 & 0.113 $\to$ 0.068 & $3\cdot10^{-6}$ \\
DE-guided MCTS             & 0.56 $\to$ 0.41 & 0.116 $\to$ 0.064 & $9\cdot10^{-6}$ \\
EDL-guided Value-Gr.       & 0.82 $\to$ 0.41 & 0.099 $\to$ 0.061 & $8\cdot10^{-8}$ \\
MCD-guided RH Orient.      & 0.65 $\to$ 0.44 & 0.062 $\to$ 0.045 & 0.008 \\
GP-guided MCTS             & 0.24 $\to$ 0.24 & 0.042 $\to$ 0.045 & 0.57 \\
$\varepsilon$-Greedy, deep   & $\sim$0.30 $\to$ $\sim$0.27 & unchanged & 0.17-0.96 \\
\bottomrule
\end{tabular}
\end{table}

\subsection{Computation time}\label{sec:res_cost}

The GP needs seconds per step and the deep models milliseconds. Over the missions, the GP takes 2.1\,s per step on average with MCTS and 2.8-4.3\,s with the other planners. To compare the models on equal inputs, we pass the same 120 logged observation sets through each model, sequentially in a single process (Table~\ref{tab:time}). The cost of the GP grows with the number of observed cells, from 0.12\,s with fewer than 100 cells to 10.8\,s with 800-1100, whereas the deep models run at a constant cost. At the median, EDL is about 700 times faster than the GP, 4 times faster than the Deep Ensembles and 35 times faster than MC Dropout.
Regarding planners, the greedy variants run in under 2\,ms, while RH Orienteering (1.3\,s) and MCTS (2.5\,s) remain compatible with ASV operation. 

Training is a one-off, offline cost. With 50 epochs on the 1600 training spills, EDL and MC Dropout train in 1.7\,min each on one RTX 3090, and the Deep Ensembles in 8.0\,min, since they train five networks. The GP has no training stage, but it pays an equivalent cost online: it refits its kernel hyperparameters at every step of every mission.

\begin{table}[t]
\centering
\footnotesize
\caption{Prediction time per step on the same 120 observation sets (deep models on GPU, GP on CPU) and training time (50 epochs, 1600 spills, RTX 3090).}
\label{tab:time}
\setlength{\tabcolsep}{4pt}
\begin{tabular}{lccc}
\toprule
Model & Mean (ms) & Median (ms) & Train (min) \\
\midrule
EDL            & \textbf{4.4} & \textbf{3.9} & \textbf{1.7} \\
Deep Ensembles & 18.5  & 16.2  & 8.0 \\
MC Dropout     & 136.3 & 137.3 & \textbf{1.7} \\
GP             & 4215  & 2851  & -- \\
\bottomrule
\end{tabular}
\end{table}

\section{Conclusion} \label{conclusion}

This paper studied how the uncertainty estimates of deep reconstruction models condition IPP for a fleet of ASVs monitoring an oil spill. MC Dropout, Deep Ensembles and EDL were used as uncertainty quantification paradigms and compared against a GP in closed loop with five planners.

Within the simulated conditions evaluated, EDL was the best model: the most accurate and the best at delineating the spill under every planner, the best calibrated in both UCE and calibration curves, and the fastest at inference. It reached an RMSE of 0.029 within the first 20\% of the budget, below the lowest error that any other model reached during the mission, and with 50 cells per vehicle it outperformed MC Dropout and Deep Ensembles with 150. Among the deep models, the choice of model explained 80\% of the variance of the final error and the planner about 1\%. No planner was best in RMSE for every model, while MCTS and RH Orienteering delineated the spill better than the greedy planners. Uncertainty-guided sampling concentrated the observations on the spill, beyond the training distribution; the Deep Ensembles failed under this shift and became more confident as their error grew, while EDL did not. The GP was the least accurate model, and its uncertainty located its error worst. When every fleet's data were reconstructed by the same model, the three deep models guided equally well with look-ahead planners, whereas greedy planners passed differences in guidance quality on to the data.



The planners used here normalize the uncertainty map and therefore only exploit where the uncertainty is high, not how large it is. Future work will study planners that treat budget and uncertainty as a multi-objective problem and use the calibrated magnitude of the uncertainty, for instance to stop sampling a region once its expected error is low enough or to allocate the remaining budget among agents. We will also consider dynamic scenarios in which the pollutant evolves as agents explore, fleets of different size, and vehicles with heterogeneous sensors, communication constraints and field validation.


 
\bibliographystyle{IEEEtran}
\bibliography{references}


\vfill

\end{document}